\documentclass[10pt,twocolumn,letterpaper]{article}

\usepackage[pagenumbers]{cvpr} % To force page numbers, e.g. for an arXiv version

\usepackage{graphicx}
\usepackage{amsmath}
\usepackage{amssymb}
\usepackage{booktabs}
\usepackage{times}
\usepackage{pifont}
\usepackage{soul}
\usepackage{url}
\usepackage{makecell}
\usepackage{multirow}
\usepackage[utf8]{inputenc}
\usepackage[pagebackref,breaklinks,colorlinks]{hyperref}
\usepackage[capitalize]{cleveref}
\usepackage{etoolbox}

\makeatletter
\patchcmd{\@maketitle}{\raggedright}{\centering}{}{}
\patchcmd{\@maketitle}{\raggedright}{\centering}{}{}
\makeatother
\begin{document}

%%%%%%%%% TITLE - PLEASE UPDATE
\title{CL-CrossVQA: A Continual Learning Benchmark for Cross-Domain Visual Question Answering}

\author{Yao Zhang \textsuperscript{\rm 1},
\qquad Haokun Chen \textsuperscript{\rm 1,4},
\qquad Ahmed Frikha \textsuperscript{\rm 1,4},
\qquad Yezi Yang \textsuperscript{\rm 2},\\
\qquad Denis Krompass \textsuperscript{\rm 1,4},
\qquad Gengyuan Zhang \textsuperscript{\rm 1},
\qquad Jindong Gu \textsuperscript{\rm 3},
\qquad Volker Tresp \textsuperscript{\rm 1,4}\\
	% Affiliations
    \textsuperscript{\rm 1} Institute of Informatics, LMU Munich \ \ \ \  
    \textsuperscript{\rm 2} Department of Informatics, Technical University of Munich \\
    \textsuperscript{\rm 3} Torr Vision Group, University of Oxford  \ \ \ \  
    \textsuperscript{\rm 4} Corporate Technology, Siemens AG \\
   	yzhang@dbs.ifi.lmu.de, \qquad volker.tresp@siemens.com
}

\maketitle

%%%%%%%%% ABSTRACT
\begin{abstract} 
   
Visual Question Answering (VQA) is a multi-discipline research task. To produce the right answer, it requires an understanding of the visual content of images, the natural language questions, as well as commonsense reasoning over the information contained in the image and world knowledge. Recently, large-scale Vision-and-Language Pre-trained Models (VLPMs) have been the mainstream approach to VQA tasks due to their superior performance. The standard practice is to fine-tune large-scale VLPMs pre-trained on huge general-domain datasets using the domain-specific VQA datasets. However, in reality, the application domain can change over time, necessitating VLPMs to continually learn and adapt to new domains without forgetting previously acquired knowledge. Most existing continual learning (CL) research concentrates on unimodal tasks, whereas a more practical application scenario, \ie, CL on cross-domain VQA, has not been studied. Motivated by this, we introduce \textbf{CL-CrossVQA}, a rigorous \textbf{C}ontinual \textbf{L}earning benchmark for \textbf{Cross}-domain \textbf{V}isual \textbf{Q}uestion \textbf{A}nswering, through which we conduct extensive experiments on 4 VLPMs, 4 CL approaches, and 5 VQA datasets from different domains. In addition, by probing the forgetting phenomenon of the intermediate layers, we provide insights into how model architecture affects CL performance, why CL approaches can help mitigate forgetting in VLPMs to some extent, and how to design CL approaches suitable for VLPMs in this challenging continual learning environment. To facilitate future work on CL for cross-domain VQA, we will release our datasets and code.

\end{abstract}

%%%%%%%%% BODY TEXT
\section{Introduction}
\label{intro}

\begin{figure}
    \centering
    \includegraphics[width =0.49\textwidth]{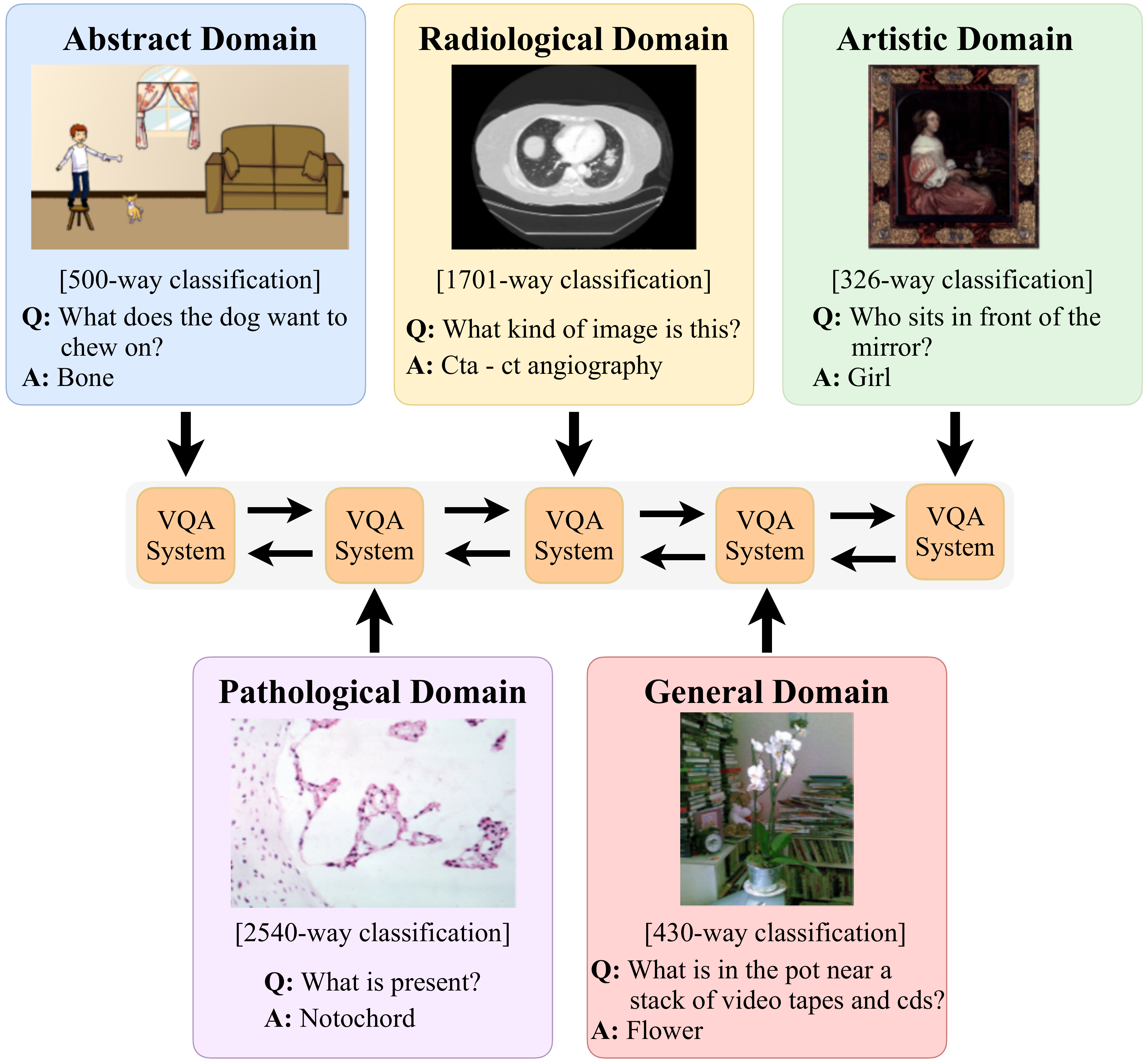}
    \caption{Illustration of CL-CrossVQA benchmark. Each VQA task is framed as a classification problem. The VQA system is trained on successive tasks in different domains and evaluated on all encountered tasks. The number of candidate answers, \ie, classes, is shown between the image and the question.}
    \label{fig:main_illu}
    \vspace{-15pt}
\end{figure}

Visual Question Answering (VQA) \cite{antol2015vqa} is an exhilarating reasoning task at the crossroads of Natural Language Processing (NLP) and Computer Vision (CV), with the objective of building a system that responds to natural language inquiries regarding an image. The challenge of VQA stems from the necessity to comprehend the semantic information in the textual and visual channels and their interplay. 

Large-scale Vision-and-Language Pre-trained Models (VLPMs) have been shown to be a viable part of modern VQA systems, as it yields state-of-the-art results by fine-tuning a single VQA dataset. In reality, however, the application scenario changes over time, requiring the VQA system to cope with non-stationary data, \ie, data from new domains become available over time. A trivial workaround is to repeat the training process each time a new domain is encountered. However, training large VLPMs is computationally intensive and requires a vast amount of training data, which might not be available in the application domain of interest. Moreover, the rapidly growing model size of VLPMs makes it impractical to store a new model for each incoming task. Thus, VLPMs are required to continually learn new concepts without forgetting previously acquired knowledge. Concretely, VLPMs need to maintain a balance between plasticity, \ie, the ability to acquire new knowledge, and stability, \ie, the ability to retain prior knowledge, which is known as the \textit{stability-plasticity dilemma} \cite{mermillod2013stability}.

In contrast to CV and NLP, CL is still underexplored in the field of Vision-and-Language (V-L) learning. While a few works \cite{greco2019psycholinguistics, lei2022symbolic} address VQA in a continual learning environment, they focus on single-domain VQA tasks. However, the more practical application scenario, \ie, CL on cross-domain VQA, has not been studied yet. 

To systematically investigate how large-scale VLPMs learn a stream of cross-domain VQA tasks, we construct \textbf{CL-CrossVQA}, a \textbf{C}ontinual \textbf{L}earning benchmark for \textbf{Cross}-domain \textbf{V}isual \textbf{Q}uestion \textbf{A}nswering, which covers 4 large-scale VLPMs, 4 CL approaches, and 5 cross-domain VQA datasets. Through comprehensive analysis of a variety of VLPMs and CL methods, we investigate how model design along multiple dimensions, \eg, text encoder, vision encoder, multimodal fusion module, affects CL performance, and which type of CL approach is more effective for VLPMs. Additionally, by dissecting the inner architecture of VLPMs using intermediate representation probing, we provide insights into
why CL approaches can help mitigate forgetting in VLPMs to some extent, and how to design CL approaches suitable for VLPMs in this challenging, yet practical continual learning environment.

Our key observations are as follows: (1) All VLPMs suffer from severe forgetting. Dual-stream encoder-decoder architecture exhibits alleviation of forgetting compared to single-stream encoder-only models; (2) For single-stream encoder-only models, the deeper layers are most affected by forgetting. For the dual-stream encoder-decoder model, the multimodal fusion module is less affected by forgetting; (3) Different VLPMs require different CL approaches to tackle the \textit{stability-plasticity dilemma}. Replay-based approaches are the most effective for VLPMs and are less sensitive to task orders; (4) For replay-based approaches, the sampling strategy used for the replay buffer has a great impact on CL performance; (5) Adapter \cite{houlsby2019parameter} is a strong competitor in this cross-domain multimodal continual learning scenario.

\section{Related Work}
\label{RW}
Our work lies at the intersection of VLPMs, VQA, cross-domain learning, and CL. We provide here a compendious overview of the relevant background. 

\subsection{Vision-and-Language Pre-trained Models}
  
In recent years, pre-trained models have progressed at a dizzying speed in tandem with the evolution of transformers. Due to the significance of single-modal language/vision pre-training, some pioneering works \cite{tan2019lxmert, lu2019vilbert, chen2019uniter, su2019vl,  li2019visualbert, wang2021simvlm} have recently attempted to explore the joint representation of language and vision by pre-training large-scale models on vision and language modalities, which are referred to as Vision-and-Language Pre-trained Models (VLPMs). There are three primary components in VLPMs, namely vision encoder (VE), text encoder (TE), and modality fusion module (MF). VE and TE are pre-trained with images and texts, respectively. MF is pre-trained using image-text pairs and amalgamates the output embeddings of VE and TE. While almost all VLPMs extract text embeddings from a pre-trained language model, \eg, BERT \cite{devlin2018bert}, the pre-trained model used on the vision side is different. Unlike most existing VLPMs, which are built upon region-level features extracted from pre-trained object detectors, \eg, Faster RCNN \cite{ren2015faster}. All VLPMs evaluated in our work directly utilize patch features without bounding box annotations, which allows them to escape from capacity limitations due to the imperfections of the object detectors.
\begin{table}
 \scalebox{0.65}{
  \centering
  \begin{tabular}{c|ccccc}
    \toprule
    \multirow{2}{*} {Model} & Vision & Text & Modality Fusion & \multirow{2}{*}{Decoder} & Pre-training \\ 
    & Encoder & Encoder& Module & &Dataset Size \\
    \midrule
    ViLT \cite{kim2021vilt} & Patch Emb. & Emb. & Single-stream & \ding{55} & 5M\\
    VAuLT \cite{chochlakis2022vault} & Patch Emb. & BERT & Single-stream & \ding{55} & 5M\\
    FLAVA \cite{singh2022flava} & ViT & ViT & Single-stream & \ding{55} & 70M \\
    ALBEF \cite{li2021align} & ViT & BERT & Dual-stream & \ding{51} & 14M\\
    \bottomrule
  \end{tabular}
}
\caption{Comparison of evaluated VLPMs.}
\label{table: model}
\vspace{-10pt}
\end{table}

To thoroughly investigate the effects of model design in a CL setting, we evaluate 4 different types of VLPMs, which are summarized in \cref{table: model}. Specifically, \textbf{ViLT} \cite{kim2021vilt} directly feeds image patch features and text token embeddings into a pre-trained ViT \cite{dosovitskiy2020image} model that learns the vision-language alignment with self-attention across both modalities. To address the impoverished language representations of ViLT, \textbf{VAuLT} \cite{chochlakis2022vault}, an extension of ViLT, propagates the output representations of a large language model, \ie, BERT, to the language input of ViLT. \textbf{FLAVA} \cite{singh2022flava}, a foundational language and vision alignment model, learns strong representations through joint pre-training on both unimodal and multimodal data while encompassing cross-modal alignment objectives and multimodal fusion objectives. FLAVA adopts the ViT architecture for the visual encoder, text encoder, as well as multimodal encoder. Unlike the aforementioned VLPMs, FLAVA is pre-trained on Public Multimodal Datasets \cite{singh2022flava}, which consist of 70M image-text pairs. As opposed to all aforementioned single-stream VLPMs, which learn one joint V+L representation, \textbf{ALBEF} \cite{li2021align} adopts dual-stream architecture to map vision and language embeddings into the same semantic space. Specifically, ALBEF employs two separate transformers for images, \ie, ViT, and texts, \ie, BERT, and introduces a contrastive loss to align the image and text representation before fusing them through cross-modal attention. Furthermore, all aforementioned VLPMs adopt the encoder-only architecture, where the cross-modal representations are directly fed into an output layer to generate the final outputs. ALBEF, on the other hand, utilizes an encoder-decoder architecture, where cross-modal representations are first fed into a decoder and then into an output layer. 
  
\subsection{Cross-Domain Learning in VQA} 
The cross-domain learning remains underinvestigated in VQA, only a few works research in this direction: \cite{chao2018cross} found that domain shift across VQA datasets mostly lies in their questions and answers and proposed an algorithm to align the source and target feature embeddings. \cite{xu2019open} trained an additional domain discriminator to adversarially penalize the mismatch between the multi-modal embedding from the source domain and the target domain. \cite{zhang2021domain} focused on domain shift in the visual space and proposed a domain-adaptive visual feature extractor to better capture image semantics from different domains. \cite{cascante2022simvqa} generated synthetic images and incorporated them in the training by  object-level feature swapping across domains.

\subsection{Continual Learning}
The main goal of continual learning research is to mitigate the problem of catastrophic forgetting (CF), \ie, the tendency of neural networks to forget existing knowledge when learning new tasks with novel input patterns. To tackle this problem, numerous studies have been conducted in different directions.

\paragraph{Rehearsal-based Methods} To mitigate forgetting, Replay methods \cite{rebuffi2017icarl, rolnick2019experience, isele2018selective, chaudhry2019continual} store and replay samples from previous tasks when learning new tasks. Dark Experience Replay (DER and DERPP) \cite{buzzega2020dark} expands ER \cite{rolnick2019experience} by distilling the dark knowledge from models on previous tasks using replayed samples. Gradient Episodic Memory (GEM) \cite{lopez2017gradient} and its lightweight variant A-GEM \cite{chaudhry2018efficient} constrain the model to update on new tasks without interfering with the previous tasks.  Instead of storing samples from previous tasks, pseudo replay methods \cite{shin2017continual, lavda2018continual, atkinson2018pseudo} optimize a generative network to produce synthetic samples depicting the knowledge of previous tasks. 

\paragraph{Regularization-based Methods} Regularization-based methods focus on estimating a distribution over the model parameters, and use prior when learning new tasks. Elastic Weight Consolidation (EWC) \cite{kirkpatrick2017overcoming} is the first to estimate the importance of network parameters and penalize parameter drifting accordingly when learning new tasks. Instead of measuring the parameter importance after each task, Synaptic Intelligence (SI) \cite{zenke2017continual} extends EWC by maintaining an online importance estimation. There are also methods that obtain priors by conducting unsupervised estimation \cite{aljundi2018memory}, knowledge distillation \cite{li2017learning, jung2016less, zhang2020class} or incrementally merging Gaussian posteriors for the task parameters \cite{lee2017overcoming}.
\paragraph{Multimodal Continual Learning Benchmarks}
Existing CL on VQA works primarily consider answer- and question-type incremental learning or scene- and function-incremental learning. For instance, \cite{greco2019psycholinguistics} divides CLEVR \cite{johnson2017clevr} into Wh-question and polar question, and examines the effect of task difficulty on CL, and if the sequence in which a kid obtains question types is beneficial to computational models. \cite{lei2022symbolic} reorganizes the existing VQA dataset to construct CLOVE, which contains two CL setting \ie, scene-incremental and function incremental, to test the ability of the model to adapt to new scenes (\eg ShopAndDinning, Workplace \etc) and acquire new features (\eg object recognition, attribute recognition \etc). In addition to these VQA benchmarks, a multimodal multi-task CL benchmark CLiMB\cite{srinivasan2022climb} has recently been introduced. In this benchmark, VLPMs are trained on a sequence of different V+L tasks.
In this paper, we consider a more practical CL setting, in which a VQA system is required to continually adapt to new domains.

\section{CL-CrossVQA Benchmark}
We first illustrate the CL-CrossVQA benchmark, then we introduce the metrics to measure the knowledge preservation and transfer abilities of different CL models.

\subsection{Task Formulation}
Given an image-question pair, the VQA system is required to answer the question based on the given image. Following previous work \cite{https://doi.org/10.48550/arxiv.1505.00468}, we consider VQA as a classification task and require the model to select a correct answer from a given answer pool, \ie, a list of answers. As mentioned before, the concurrent works investigating CL+VQA focus on single-domain VQA tasks, and CL on cross-domain VQA remains unexplored.

We formalize our goal of CL on a sequence of $K$ VQA tasks $\{T_1, ..., T_K\}$, where each task $T_k$ contains $N_k$ image-question-answer triplets $\{(v_k^i, q_k^i, a_k^i) |\ i \in \{1, .., N_k\}\}$ from a specific domain $D_k$. Here, domain $D_k$ is specified by the joint distribution of $\mathcal{P}_{VQA}^k$, where $\mathcal{P}^k_{V}$, $\mathcal{P}^k_{Q}$, $\mathcal{P}^k_{A}$ are the marginal distribution of the images, questions, and answers from that domain, respectively. We define our cross-domain learning by setting $\mathcal{P}_{VQA}^k \neq \mathcal{P}_{VQA}^j$ with $\forall j,k \in \{1, .., K\}$, where $ j \neq k$. The answer pool $A_k$ of each task $T_k$ include $C_k$ ground-truth answers, \ie, $A_k=\{a_k^1,..., a_k^{C_k}\}$. Note that, the size of the answer pool could be different for different tasks. The goal is to optimize a set of parameters $\widetilde{\Theta}$ of the learner towards an objective on each task $T_k$.

\subsection{Datasets}

CL-CrossVQA consists of 5 datasets across different domains, namely the abstract domain, \ie, VQA Abstract\cite{https://doi.org/10.48550/arxiv.1505.00468}, the general domain, \ie, Toronto COCO QA\cite{DBLP:journals/corr/RenKZ15}, the artistic domain, \ie, AQUA\cite{garcia2020AQUA}, the pathological domain, \ie, PathVQA\cite{he2020pathvqa} and the radiological domain, \ie,  VQA-Med-2019\cite{ImageCLEFVQA-Med2019}. Examples from these datasets are provided in \cref{fig:main_illu}.

\textbf{VQA Abstract} consists of images of abstract scenes.
Each image is accompanied by three questions, with 10 human-provided ground truth answers, ensuring a diverse and interesting set of questions and answers. By its abstraction, it requires high-level reasoning skills for VQA.

\textbf{COCO QA} contains questions automatically generated from captions in Microsoft COCO dataset\cite{cocodataset} and therefore may contain syntactic and semantic errors. There are in total 78,736 training questions and 38,948 test questions, of which the answers are all one word. 

\textbf{AQUA} Artwork has been an important part of human history. Different styles of artwork such as naturalism, realism, and cubism represent different levels of abstraction, posing an extra challenge for VQA system. Aiming to investigate this, AQUA is built on the SemArt dataset\cite{garcia2018how} and its QA pairs are generated from the associated comments of the paintings. We use the Visual QA subset from AQUA, including 29,568, 1,507, and 1,270 question-answer (QA) pairs for training, validation, and testing, respectively.

\textbf{PathVQA} is created with the goal to develop an AI pathologist and foster research in pathological VQA. It consists of 32,799 question-answer pairs automatically generated from captions of 4998 pathology images in pathology textbooks and online
digital libraries. We note that the answers in PathVQA can be long, clinical expressions, adding to the difficulty of the questions.

\textbf{VQA-Med-2019} aims to support clinical education, clinical decision, and patient education. It focuses on radiology images selected from the MedPix database and  the QA pairs are generated based on patterns from naturally asked and validated questions by medical students. There are 3,200 images and 12,792 QA pairs for the training set and 500 images with 2,000 QA pairs for the validation set. The test set includes 500 images with 500 questions. 

To avoid bias caused by different dataset sizes, we downsample the training sets of all datasets to be consistent with VQA-Med-2019 while maintaining their original class distributions. The same train-test split ratio is adopted for all datasets. Detailed dataset pre-processing is provided in the Appendix. To better reflect the challenging real-world scenarios, the selected cross-domain datasets depict minor overlap in the answer space, \ie, only a few answers are shared across datasets.

\subsection{Evaluation Metrics}
To evaluate the knowledge transfer ability of the models and their possible catastrophic forgetting of the previously learned tasks, we assume access to the test set of each task and report the following three evaluation metrics \cite{lopez2017gradient}: 

\begin{equation}
\label{eq:fwt}
\textbf{Forward Transfer} \quad FWT = \frac{1}{T-1} \sum_{i=1}^{T-1} S_{i,i+1} - \overline{b}_{i+1},
\nonumber
\end{equation}
\begin{equation}
\label{eq:bwt}
\textbf{Backward Transfer} \quad BWT = \frac{1}{T-1} \sum_{i=1}^{T-1} S_{T,i} - S_{i,i},
\nonumber
\end{equation}
\begin{equation}
\label{eq:acc}
\textbf{Average Accuracy} \quad Acc = \frac{1}{T} \sum_{i=1}^{T} S_{T,i},
\nonumber
\end{equation}
where $S_{i,j}$ is the evaluation score of the model on the test set of $j$-th task after training on $i$-th task, and $\overline{b}_{i}$ is the evaluation score of the pre-trained model, \ie, model without any specific fine-tuning for VQA tasks, on the test set of $i$-th task. Specifically, we feed the task-specific answer list into the decoder of pre-trained ALBEF to compute $\overline{b}_{i}$, while for the encoder-only backbones, we attach a randomly initialized classification head on the top for each task. 

\section{Experiments}
\label{exp}
We provide a comprehensive analysis of VLPMs with our CL-CrossVQA benchmark. Concretely, we investigate a CL baseline and 4 CL approaches from different categories: (1) \textbf{Sequential} uses the model learned on previous tasks as initialization and then optimizes the parameters for the current task. (2) \textbf{ER} \cite{rolnick2019experience} is a rehearsal-based method, which stores a portion of samples from the previous task and replays them regularly with the training batch of the current task. (3) \textbf{DER} \cite{buzzega2020dark} is an extension of ER, which additionally stores the model prediction of the replayed samples at the end of the previous task as dark knowledge regularization. (4) \textbf{DERPP} \cite{buzzega2020dark} combines ER and DER and is a strong hybrid method. 
(5) \textbf{EWC} \cite{kirkpatrick2017overcoming} is a regularization-based method, which penalizes the drifting of important network parameters when training on new tasks. 

To analyze the effects of VLPM architecture in the CL setting, we train 4 representative VLPMs with the aforementioned CL methods using our benchmark. To handle the distribution shift in the ground-truth answers of different tasks, we incorporate task-specific classification heads with the encoder-only backbones, \ie, \emph{ViLT, VAuLT, and FLAVA}. For the encoder-decoder learner, \ie, \emph{ALBEF}, we implement a static model architecture for all tasks and utilize different answer lists for each task.  The hyperparameters used in the training are detailed in the Appendix. 

\begin{table*}
 \scalebox{0.63}{
  \centering
  \begin{tabular}{c |ccc |ccc| ccc | ccc}
    \toprule
    \multirow{3}{*}{Model}&
    \multicolumn{3}{c}{ViLT} &
    \multicolumn{3}{c}{VAuLT} &
    \multicolumn{3}{c}{FLAVA} &
    \multicolumn{3}{c}{ALBEF}   \\
    \cmidrule(lr){2-4} 
    \cmidrule(lr){5-7} 
    \cmidrule(lr){8-10}
    \cmidrule(lr){11-13}
         & Acc & BWT & FWT & Acc & BWT & FWT & Acc & BWT & FWT & Acc & BWT & FWT \\
    \midrule
   
    Sequential & 26.82$\pm$2.29 & -42.49$\pm$3.41 & -0.05$\pm$0.06 & 26.21$\pm$4.79 &-42.58$\pm$5.64 &-0.14$\pm$0.24 & 26.61$\pm$2.78 &-34.31$\pm$2.36 &-0.02$\pm$0.40 &45.71$\pm$3.43 & -27.13$\pm$3.69 & 9.77$\pm$2.83 \\
    
	ER & 54.15$\pm$1.36 &-12.38$\pm$1.82 &\textbf{0.03$\pm$0.06 }& \textbf{51.51$\pm$0.91} &\textbf{-12.67$\pm$1.07} &0.08$\pm$0.12 & 44.52$\pm$0.80 &\textbf{-10.53$\pm$1.14 }&-0.11$\pm$0.18 & \textbf{60.79$\pm$0.54} &\textbf{-9.77$\pm$1.02} &11.90$\pm$2.39\\
	
	DER & 51.42$\pm$1.71 &-12.56$\pm$1.58 &-0.15$\pm$0.38 & 49.35$\pm$1.29 &-14.86$\pm$1.63 &-0.15$\pm$0.41 & \textbf{44.82$\pm$1.09} &-10.91$\pm$2.75 &-0.07$\pm$0.16 & 51.49$\pm$2.08 &-21.18$\pm$2.29 &\textbf{12.48$\pm$1.26}\\
 
	DERPP & \textbf{54.21$\pm$1.31} &\textbf{-12.34$\pm$1.70} &0.00$\pm$0.28 & 51.30$\pm$1.03 &-13.08$\pm$1.12 &\textbf{0.23$\pm$0.31} & 44.52$\pm$1.57 &-11.30$\pm$2.64 &-0.06$\pm$0.33 & 59.84$\pm$0.97 &-10.89$\pm$1.27 & 11.91$\pm$3.14\\
 
    EWC & 26.94$\pm$2.75 &-42.86$\pm$3.96 &0.03$\pm$0.06 & 25.43$\pm$3.79 &-43.64$\pm$4.02 &-0.03$\pm$0.60 & 25.83$\pm$4.89 &-34.09$\pm$5.87 &\textbf{-0.01$\pm$0.18} & 46.57$\pm$5.82 &-27.67$\pm$7.07 &9.62$\pm$3.53\\
    
    \bottomrule
  \end{tabular}
  }
  \caption{Evaluation of selected VLPMs combined with different CL approaches. Average accuracy (Acc), backward transfer (BWT), and forward transfer (FWT) is reported. 6 task orders are adopted to produce the mean $\pm$ standard deviation. All models are fine-tuned for a maximum of 15 epochs. We apply a random sampling strategy for ER, DER, and DERPP. For all replay-based approaches, the memory size is set to 1\% of training data. We tune the weight for logits distillation and the weight for experience replay of DERPP by grid search in \{0.1, 0.5, 1, 10, 100\}. We tune the regularization coefficient of EWC by grid search in \{0.1, 1, 10, 100, 1000, 10,000, 100,000, 1,000,000\}.}
  \label{tab:result_main}
  \vspace{-10pt}
  \end{table*}

\subsection{Results and Discussion}
In \cref{tab:result_main}, we provide the evaluation results of the selected VLPMs in our CL-CrossVQA benchmark. We report three metrics, \ie, average accuracy (Acc), backward transfer (BWT), and forward transfer (FWT). All experiments are conducted with 6 randomly sampled task orders. In the following, we discuss our findings.

\paragraph{All VLPMs evaluated suffer from severe catastrophic forgetting.} Comparing the backward transfer of all CL methods, we observe that sequential fine-tuning of all VLPMs results in the largest forgetting, indicating that the selected VLPMs cannot preserve the knowledge learned on the previous tasks. Therefore, continually fine-tuning VLPMs without applying any CL approaches would result in a considerable performance loss.

\paragraph{Replay-based approaches are the most effective for VLPMs.} Comparing the four different CL methods, we can observe that replay-based CL approaches, \ie, ER, DER, and DERPP, achieve higher accuracy and higher backward transfer than the regularization-based method, \ie, EWC, for all VLPMs. ER, and DERPP, which adopts both regularization and experience replay, achieve comparable results on our benchmark. Note that, we only use 1\% of the training data, which implies that a small number of replayed data is already able to alleviate forgetting to a large extent. Despite the careful tuning of the regularization coefficient, EWC performs as poorly as the sequential fine-tuning baseline, demonstrating that EWC is not suitable for VLPMs in the cross-domain continual learning setting. 

\paragraph{Dual-stream encoder-decoder architecture is more resistant to forgetting.} ALBEF shows the highest CL performance among all VLPMs, suggesting that the dual-stream encoder-decoder model is significantly more resistant to forgetting than the encoder-only model. Note that, ALBEF allows one to perform continual learning without adding new task-specific parameters. However, for single-stream models, new classification heads have to be added each time we train on a new task since each task in our benchmark has a different number of classes. Such differences may partially explain why ALBEF is more suitable for CL. Also, ALBEF produces the highest FWT score among all backbones, which further indicates the benefit of applying the same parameters for all tasks. In order to examine whether adding task-specific parameters is responsible for the low performance of single-stream models, we take steps toward understanding why ALBEF seems to be resistant to forgetting by probing the inner layer of ALBEF and comparing it to the single-stream representative VLPM, \ie, ViLT in \cref{sec:model}. 

Compared with the dual-stream model, single-stream models benefit more from replay-based approaches. Among all single-stream models, ViLT reaches the highest accuracy and backward transfer, implying that VAuLT, which extracts rich language representations from a pre-trained frozen BERT that serves as input embeddings for ViLT, cannot reap benefits for CL. Counterintuitively, FLAVA, which is pre-trained on the largest dataset with 70M image-text pairs, shows the worst performance in our experimental setting, suggesting that a diverse large-scale pre-training dataset is not a useful ingredient in alleviating forgetting when the domain shift of different CL tasks is large. We hypothesize that large data distribution shifts between the pre-training data of FLAVA and the datasets of our benchmark impair the transfer of the rich knowledge gained from the large pre-training dataset to the new domain, especially when the data in the new domain is limited. This phenomenon suggests that ViLT already contains rich multimodal representations, and a relatively small number of parameters makes it more efficient when transferring to a new domain and better suited for CL with limited data per task.
 
\paragraph{Impact of task orders.}

To investigate how task orders affect CL performance, we evaluate 6 task orders selected at random from all possible combinations. We find that replay-based approaches are less sensitive to task orders in comparison to
\begin{figure}[ht]
\begin{subfigure}[b]{0.24\textwidth}
    \centering
    \includegraphics[scale=0.3]{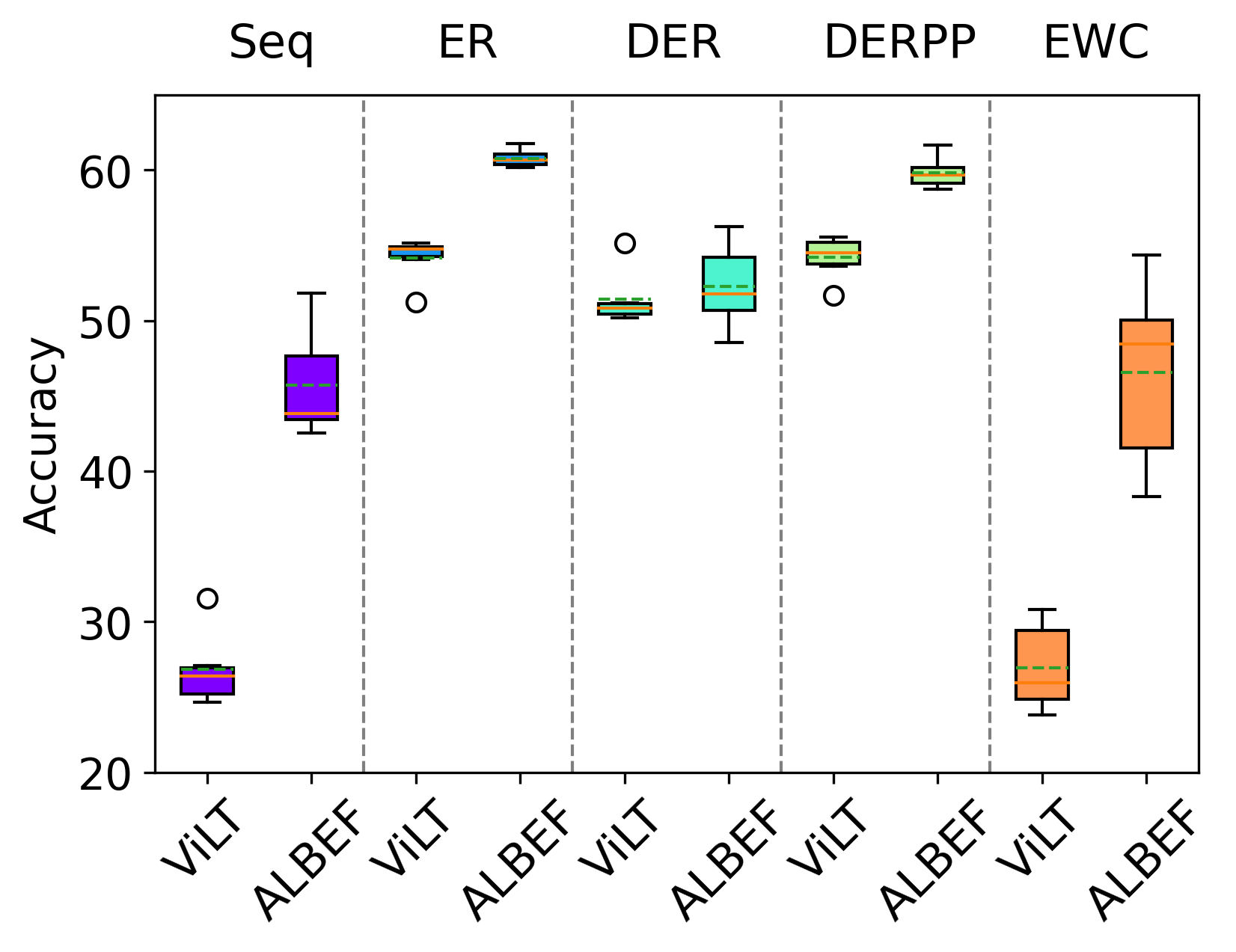}
    \caption{Acc}  
\end{subfigure}%
\begin{subfigure}[b]{0.24\textwidth}
    \centering
    \includegraphics[scale=0.3]{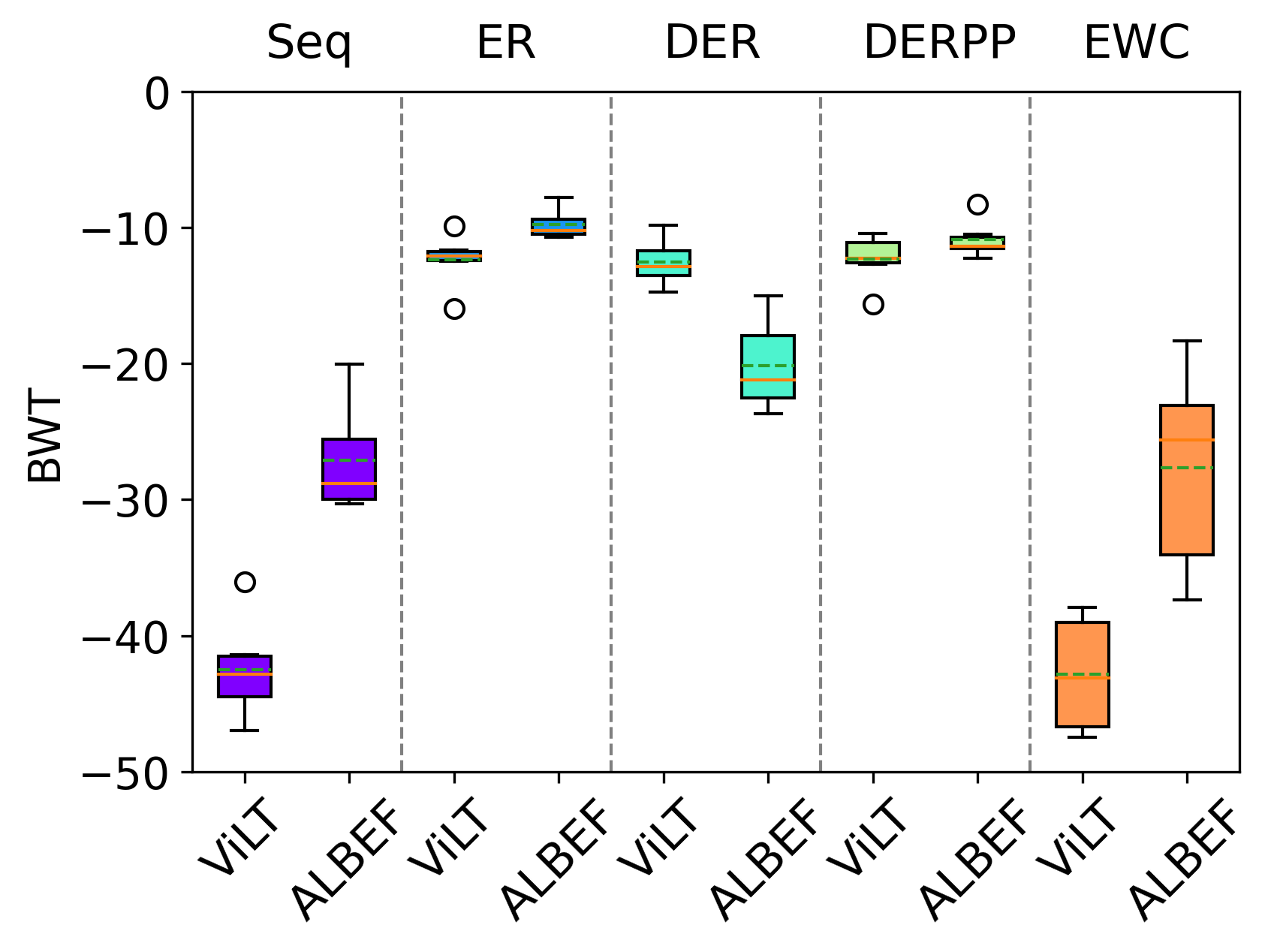} 
    \caption{BWT} 
\end{subfigure}%
\caption{Analysis of model sensitivity to task order for ViLT and ALBEF with different CL methods.}
\label{fig:task_order} 
\vspace{-10pt}
\end{figure}
regularization-based approaches (see standard errors in \cref{tab:result_main}). Among all VLPMs, ALBEF is more robust to task orders using the best-performing CL approaches, \ie, ER, and DERPP (see \cref{fig:task_order}).

\paragraph{Impact of memory buffer.}

\begin{figure}[ht]
\begin{subfigure}[b]{0.235\textwidth}
    \centering
    \includegraphics[scale=0.3]{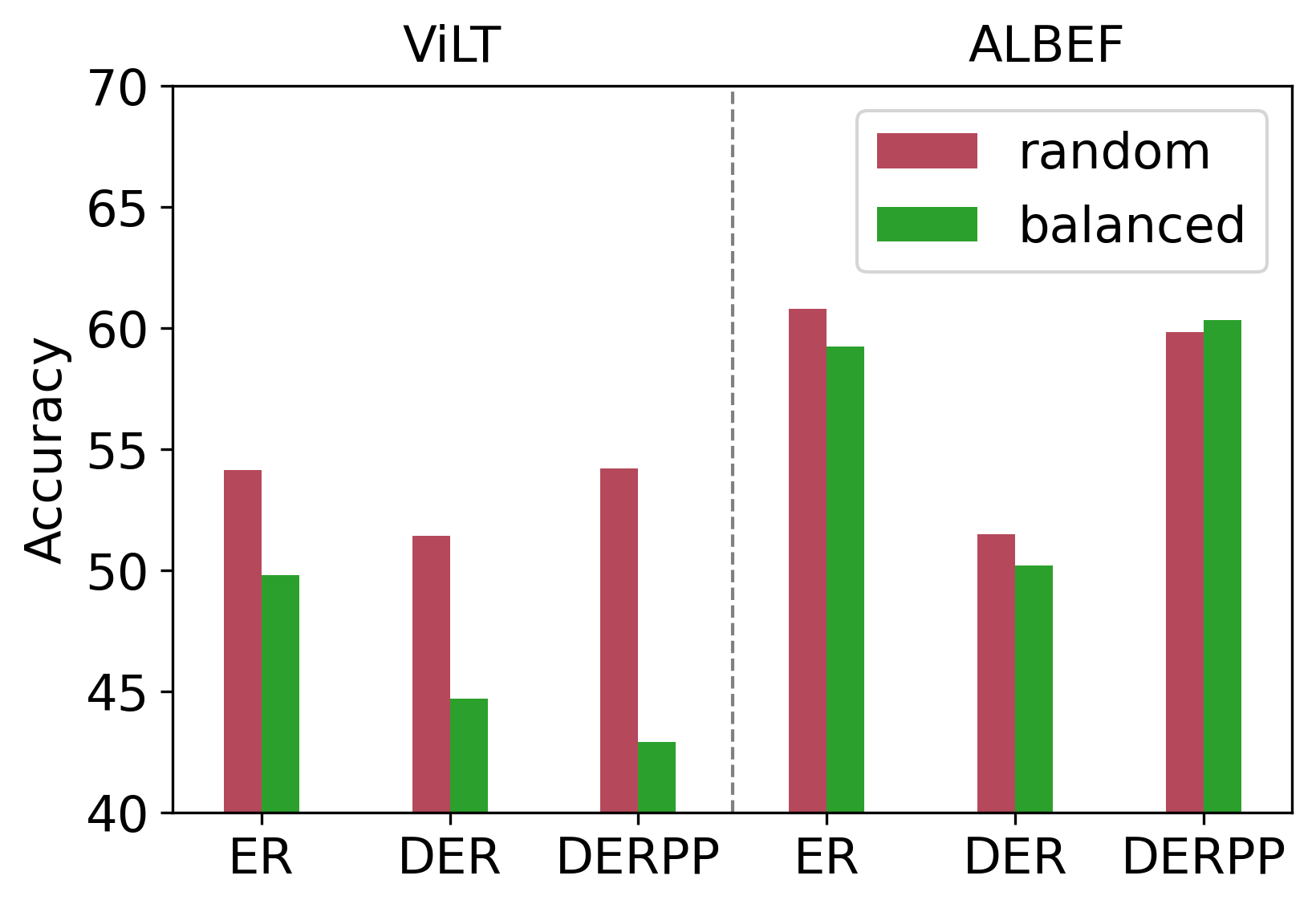} 
    \caption{Acc} 
\end{subfigure}%
\begin{subfigure}[b]{0.25\textwidth}
    \centering
    \includegraphics[scale=0.3]{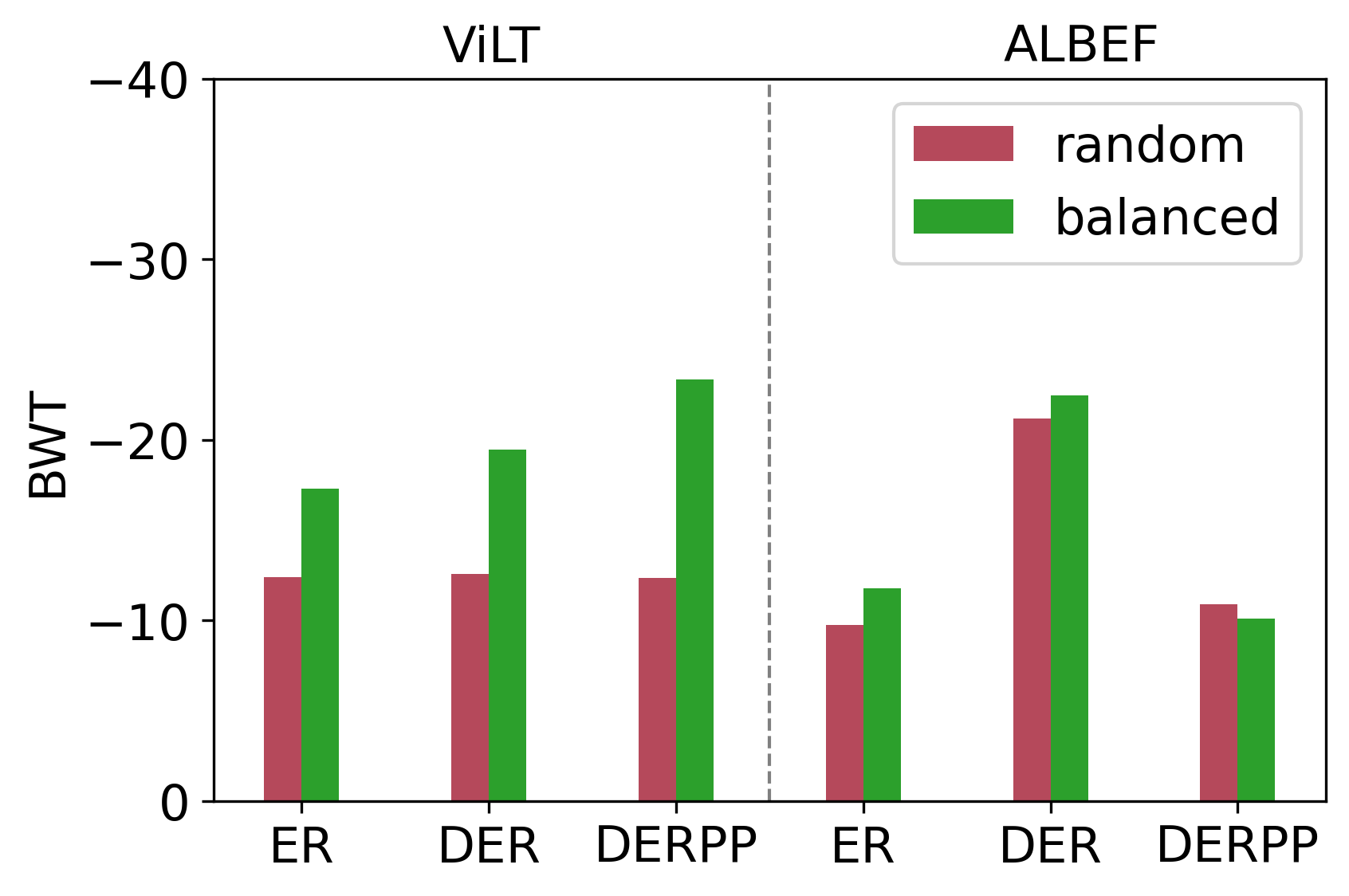} 
    \caption{BWT} 
\end{subfigure}%
\caption{Comparison of accuracy (Acc$\uparrow$) and backward transfer (BWT$\downarrow$) of replay-based approaches using different sampling strategies.}
\label{fig:balance}
\vspace{-10pt}
\end{figure}

To investigate the effect of the answer distribution in the replay buffer on the CL performance, we compare two different methods for sampling replay instances: random sampling, \ie, sampling all instances from the previous tasks with equal probability, and balanced sampling, \ie, memory buffer exhibits a class-balanced distribution regardless of the ground-truth class distribution in previous answer lists. The evaluation results are shown in \cref{fig:balance}. We observe a clear performance gap between the two sampling strategies for all 3 CL methods, indicating that replaying samples based on the class distribution of previous tasks is beneficial for continual learning. It is also important to note that this effect of the replay buffer on CL performance does not seem to be architecture-dependent. This shows that the selection of sampling strategy has a great impact on CL performance, and improving the sampling strategy could be one direction to obtain better CL performance.

\paragraph{Comparison with Adapter} In addition to conventional CL approaches, we also conduct experiments with Adapter \cite{houlsby2019parameter}, which adds new trainable modules between the intermediate layers of VLPMs. We implement the same adapter architecture for two representative VLPMs, \ie, ALBEF and ViLT. We freeze the parameters of all layers and continually optimize the task-specific adapters and the classification head for ViLT. As shown in \cref{tab:vilt_freeze} and \cref{tab:albef_freeze}, Adapter enables both models to learn new tasks without any performance degradation on the previously learned tasks, \ie, no catastrophic forgetting. Adapter is beneficial for the single-stream model as it boosts performance by ca. 11\% accuracy while being ca. 3 times more computationally efficient. In contrast, for the dual-stream model, Adapter leads to a drop of 8\% in retained accuracy compared to the best-performing CL approach. Therefore, Adapter is a strong competitor for the more computationally expensive CL methods.

\begin{table}
 \scalebox{0.94}{
  \centering
  \begin{tabular}{c|c|ccc}
    \toprule
    Method & Trainable Parameters & Acc & BWT & FWT \\
    \midrule
    DERPP & 125.97M (100.00 \%) & 54.74 & -12.22 & 0.11 \\  
	Adapter & 42.72M (33.91\%)& \textbf{65.10} & / & / \\
    \bottomrule
  \end{tabular}
  }
  \caption{Comparison of Adapter and the best-performing CL approach for ViLT.}
  \label{tab:vilt_freeze}
  \end{table}

\section{Model Analysis}
\label{sec:model}
In this section, we provide a comprehensive analysis of the impact of each individual module of VLPMs when trained in a CL environment. Specifically, we investigate the influence of vision and text encoders in \cref{sec:impact_vete}, and examine the impact of multimodal fusion designs in \cref{sec:mmfusion}. Finally, we provide some Grad-CAM visualizations for VQA after training on each task in \cref{sec:vis}. For ablation and analysis, we mainly focus on ViLT and ALBEF. The task order we use is illustrated in \cref{fig:main_illu}.

  \begin{table}
 \scalebox{0.84}{
  \centering
  \begin{tabular}{c|c|ccc}
    \toprule
    Method & Trainable Parameters & Acc & BWT & FWT \\
    \midrule
    Seq. & 290.34M (100.00\%) & 43.32 & -30.16 & 10.16\\
    ER & 290.34M (100.00\%) & \textbf{60.19} & -10.73 & 12.71\\
    Freeze VE & 204.25M (70.35\%) & 38.50 & -36.12 & 11.53\\
    Freeze TE & 223.97M (77.14\%) & 39.15 & -36.73 & 11.60 \\
        
	Freeze VE+TE & 137.88M (47.49\%) & 40.57 & -34.20 & 11.42 \\   	
	Freeze B9 & 109.53M (37.72\%)& 38.72& -34.10 & 11.35 \\ 
	Freeze B12 & 81.17M (27.96\%)& 34.69 & -33.23 & 8.46\\  
	Adapter & 85.67M (29.51\%)& 52.44 & / & /\\
    \bottomrule
  \end{tabular}
  }
  \caption{Model analysis of ALBEF with partial freeze.}
  \label{tab:albef_freeze}
  \vspace{-10pt}
  \end{table}

  \begin{figure*}[ht]
\begin{subfigure}[b]{0.333\textwidth}
    \centering
    \includegraphics[scale=0.3]{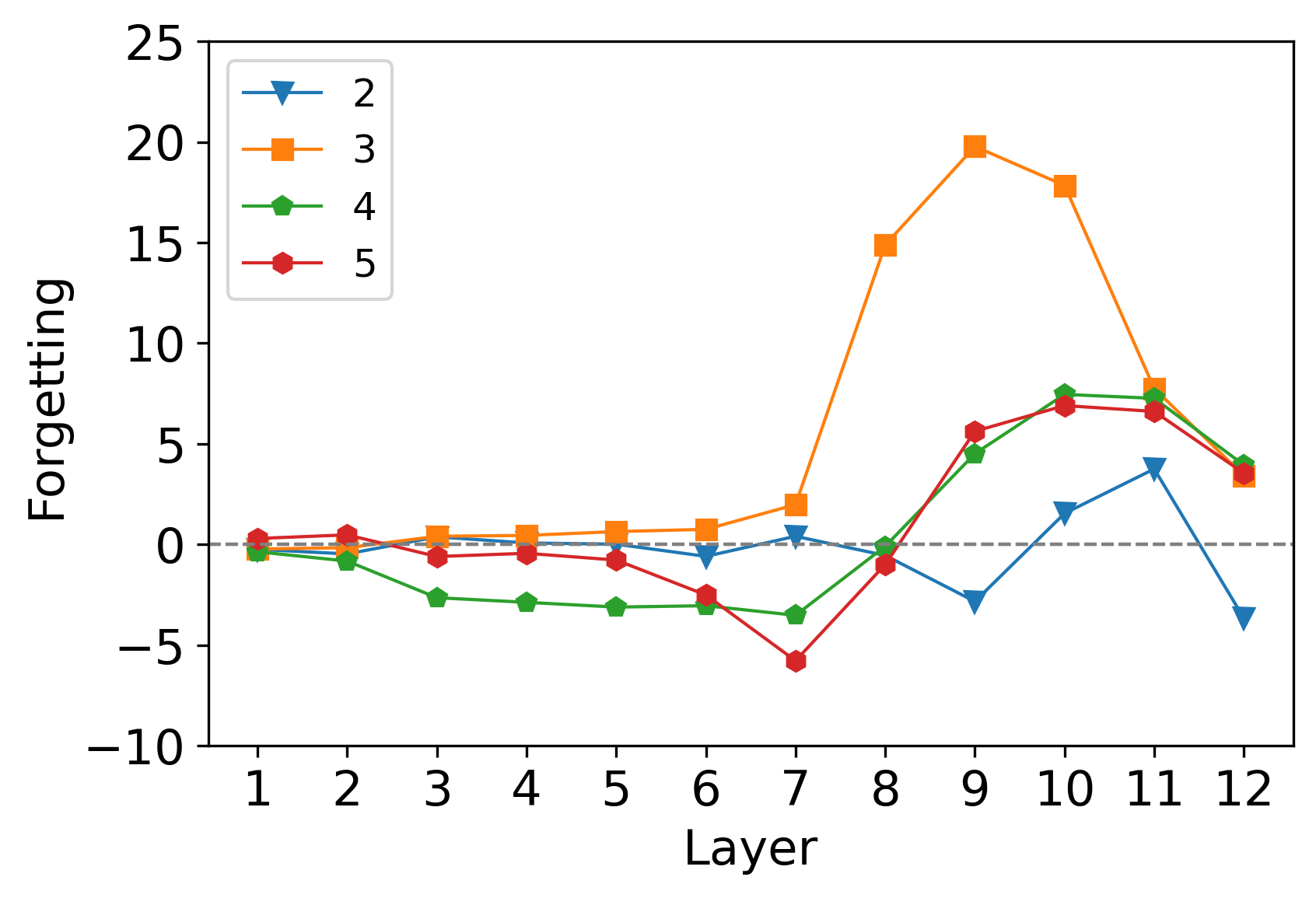} 
    \caption{ViLT + Seq} 
    \label{fig:layer_prob_vilta} 
\end{subfigure}%
\begin{subfigure}[b]{0.333\textwidth}
    \centering
    \includegraphics[scale=0.3]{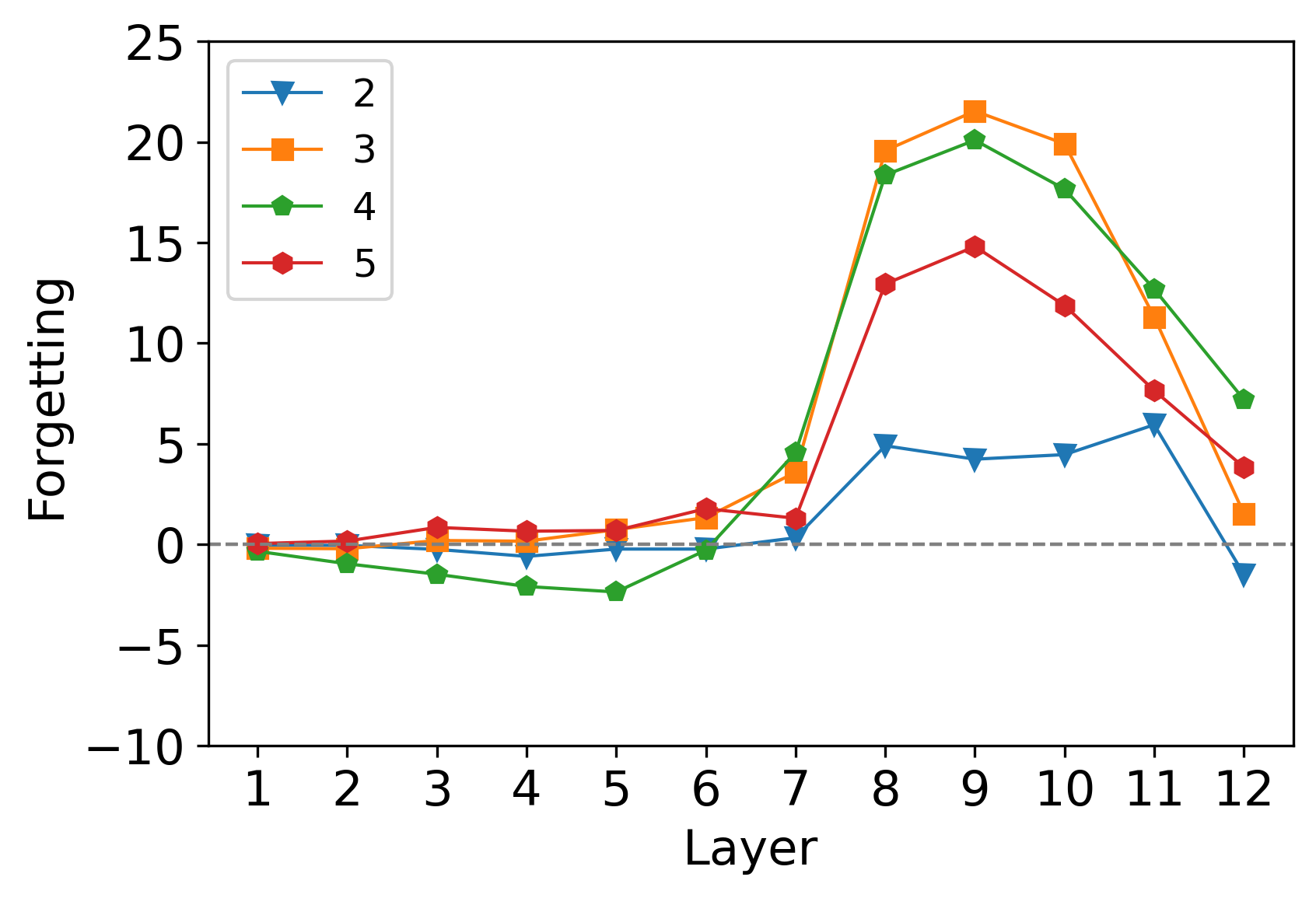} 
    \caption{ViLT + ER} 
\label{fig:layer_prob_viltb} 
\end{subfigure}%
\begin{subfigure}[b]{0.333\textwidth}
    \centering
    \includegraphics[scale=0.3]{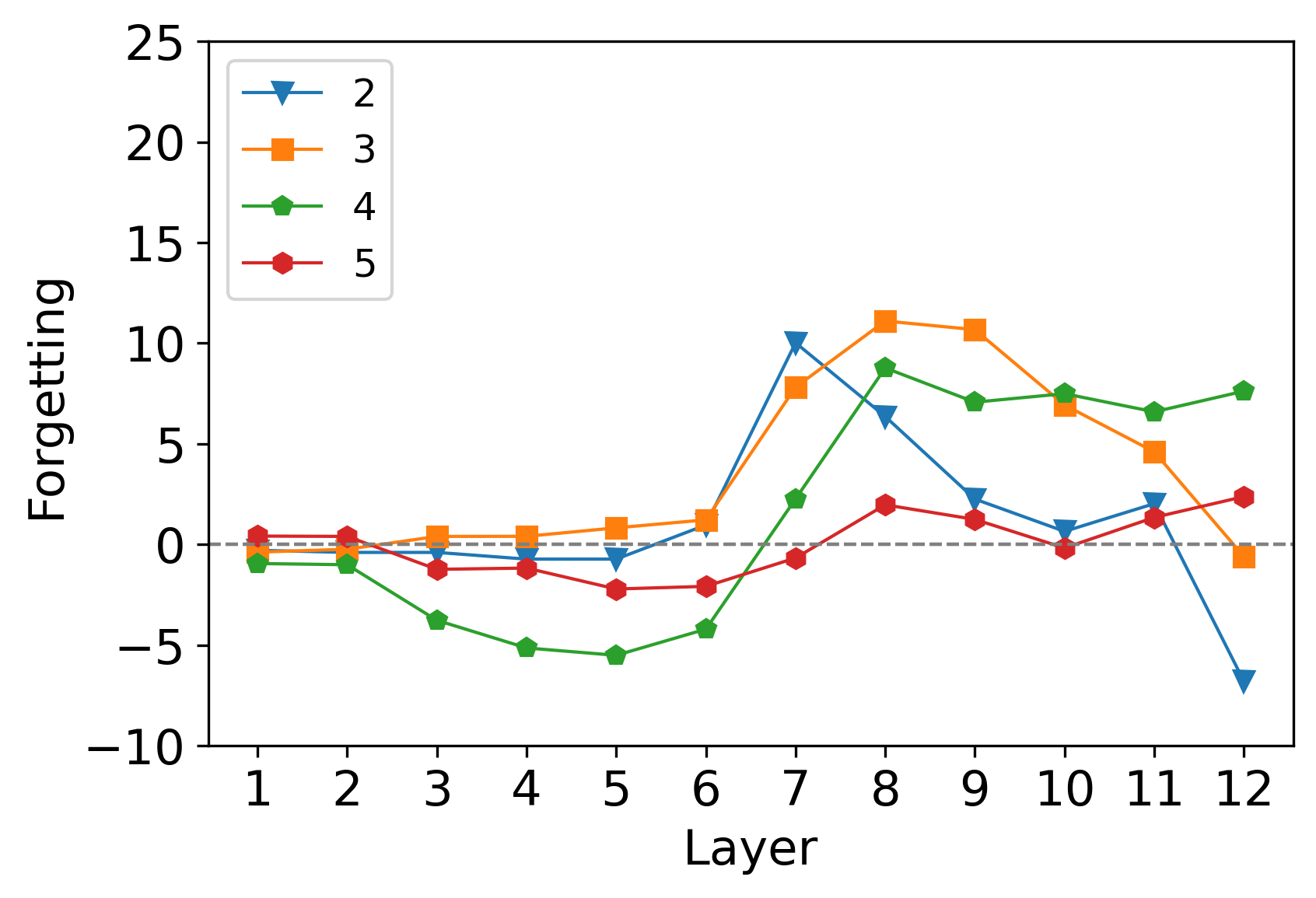} 
    \caption{ViLT + EWC} 
\label{fig:layer_prob_viltc} 
\end{subfigure}%

\begin{subfigure}[b]{0.333\textwidth}
    \centering
    \includegraphics[scale=0.3]{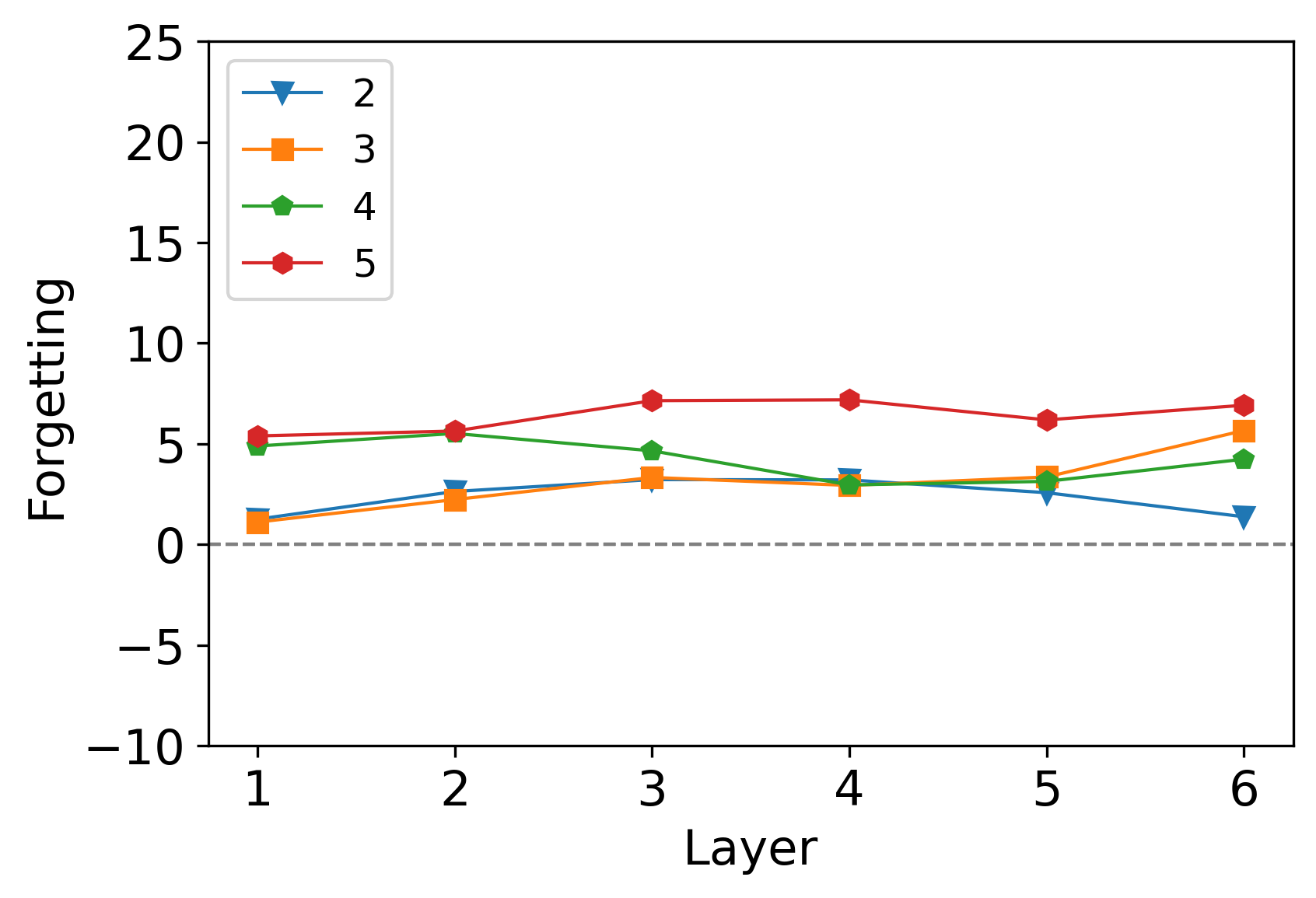} 
    \caption{ALBEF + Seq} 
\label{fig:layer_prob_viltd} 
\end{subfigure}%
\begin{subfigure}[b]{0.333\textwidth}
    \centering
    \includegraphics[scale=0.3]{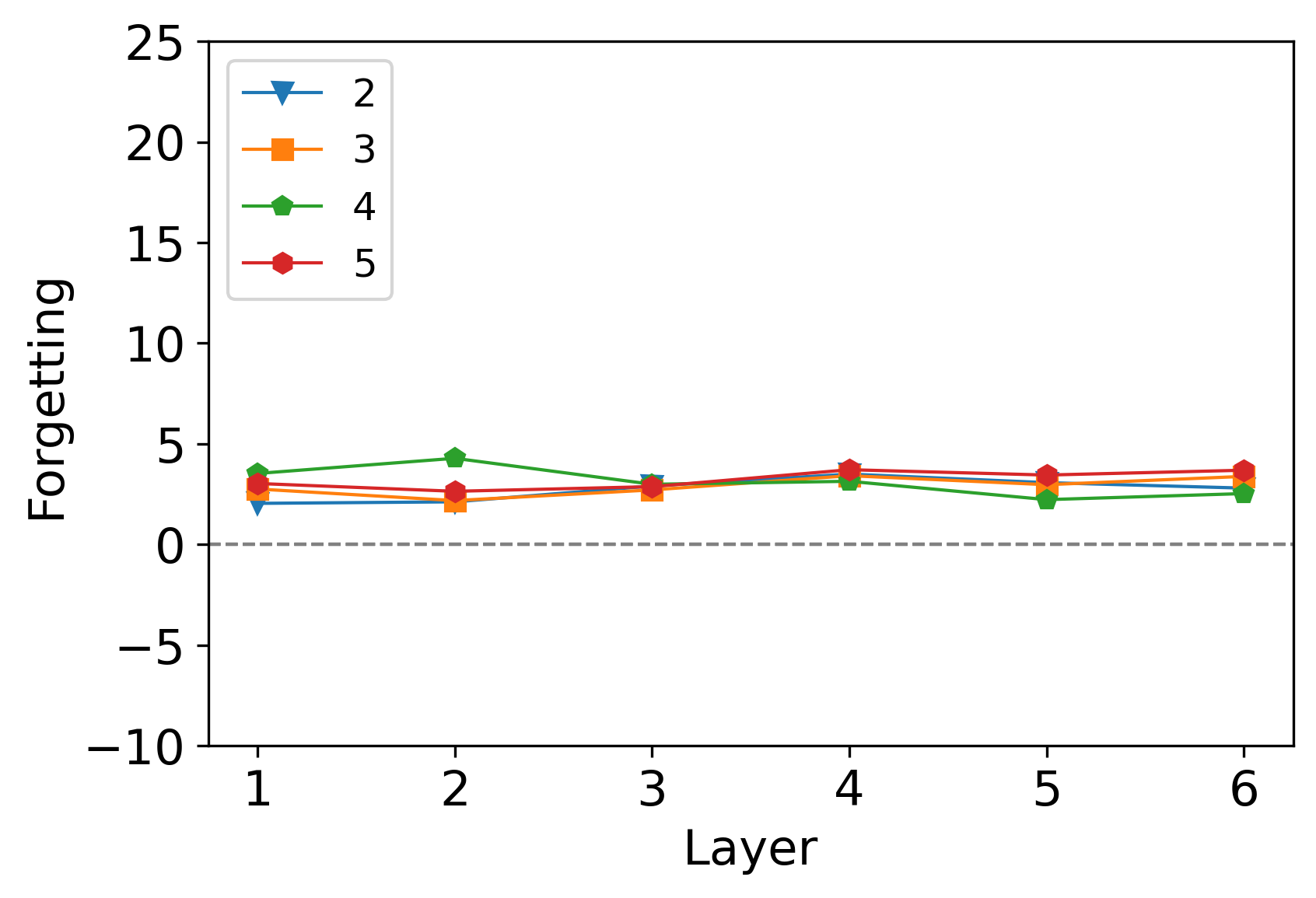} 
    \caption{ALBEF + ER} 
\label{fig:layer_prob_vilte} 
\end{subfigure}%
\begin{subfigure}[b]{0.333\textwidth}
    \centering
    \includegraphics[scale=0.3]{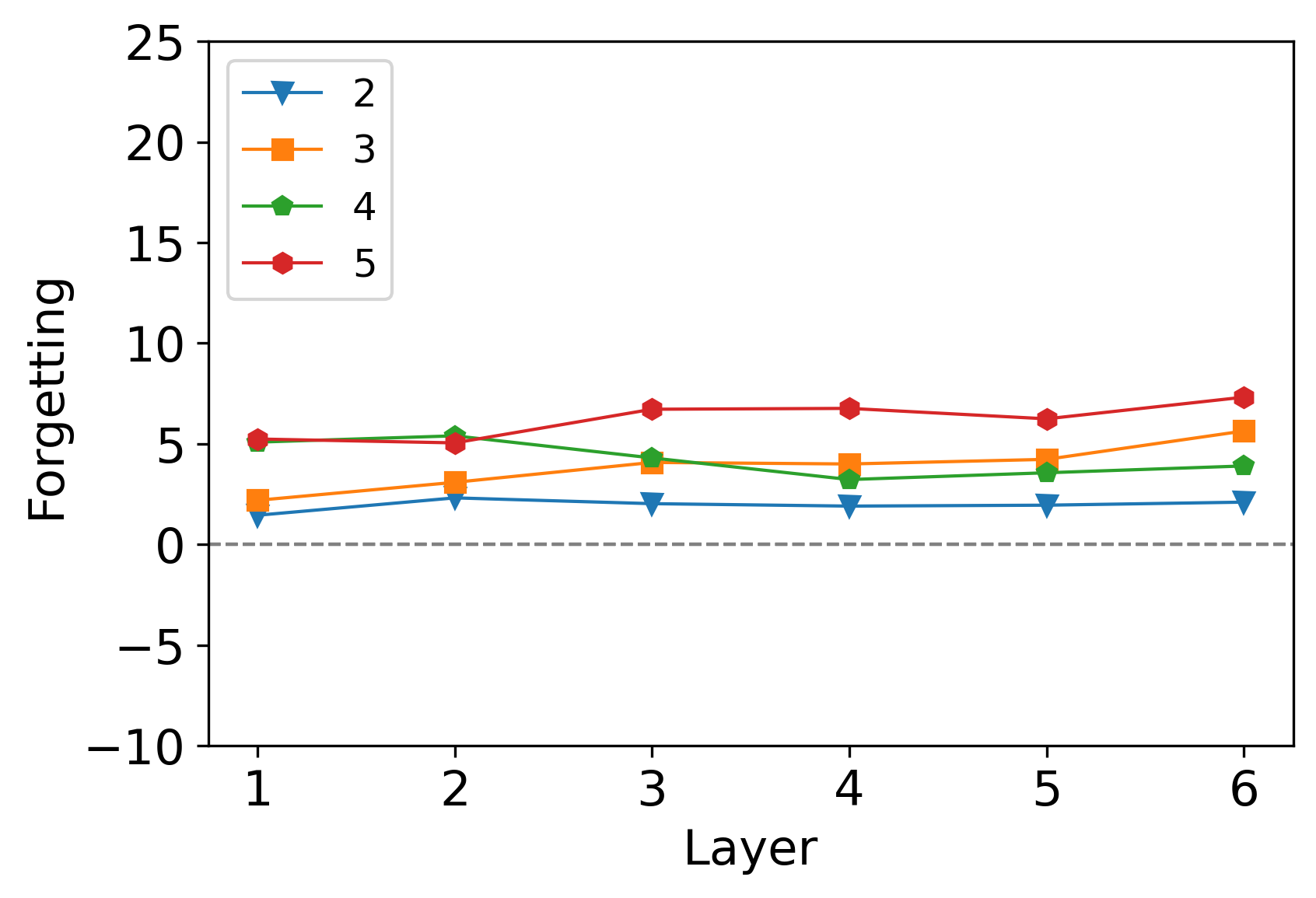} 
    \caption{ALBEF + EWC} 
\label{fig:layer_prob_viltf} 
\end{subfigure}%
\caption{Layer-wise probing results for ViLT and ALBEF with different CL methods. Each line is the \textit{representation forgetting} after the model is trained with task i. (a-c): x-axis is the layer of ViLT encoder. (d-e): x-axis is the layer of the multimodal encoder of ALBEF.}
\label{fig:layer_prob_vilt} 
\end{figure*}

\begin{figure*}[ht]
\begin{subfigure}[b]{0.25\textwidth}
    \centering
    \includegraphics[scale=0.3]{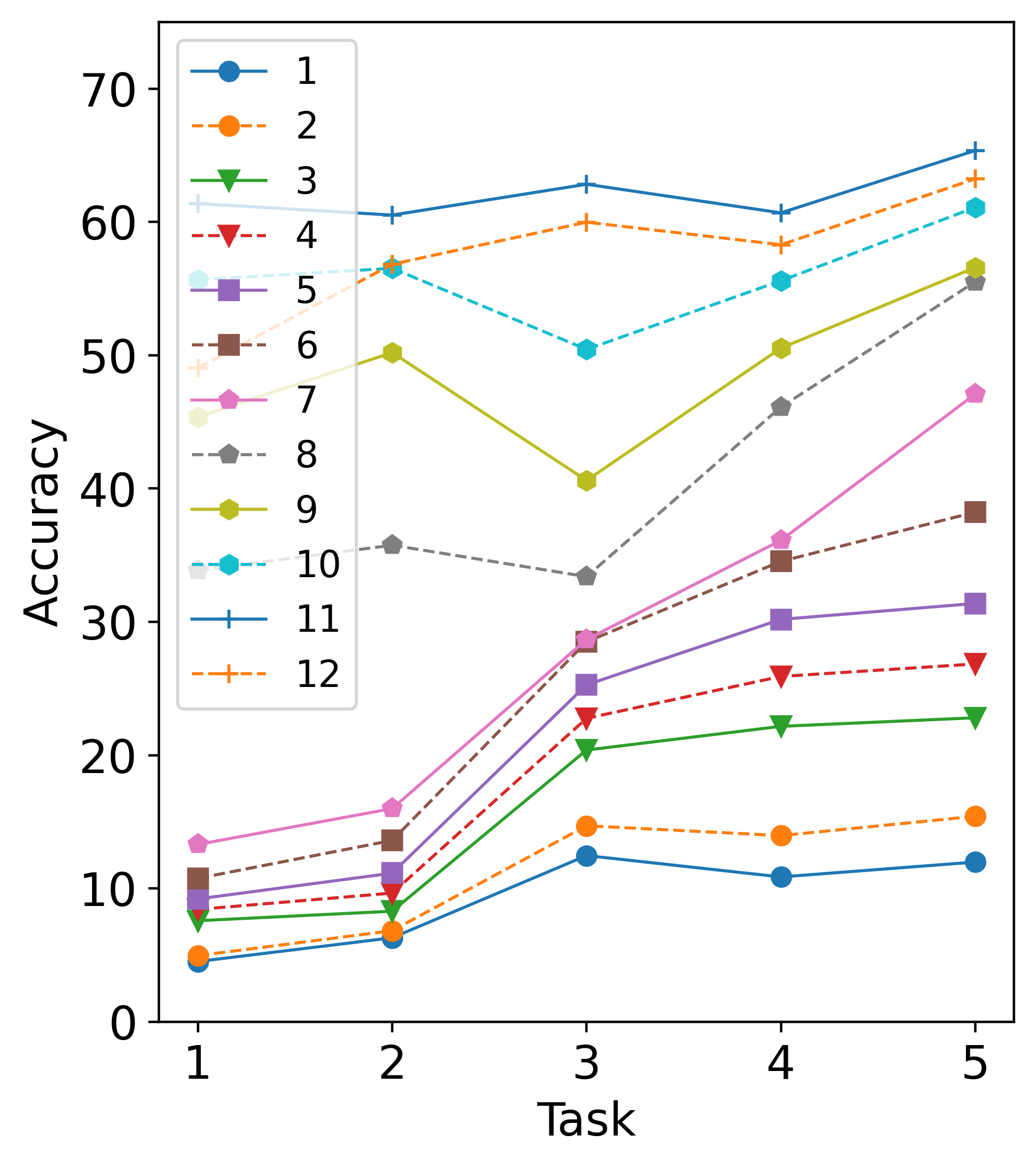} 
    \caption{ViLT + Seq} 
\label{fig:task_probinga} 
\end{subfigure}%
\begin{subfigure}[b]{0.25\textwidth}
    \centering
    \includegraphics[scale=0.3]{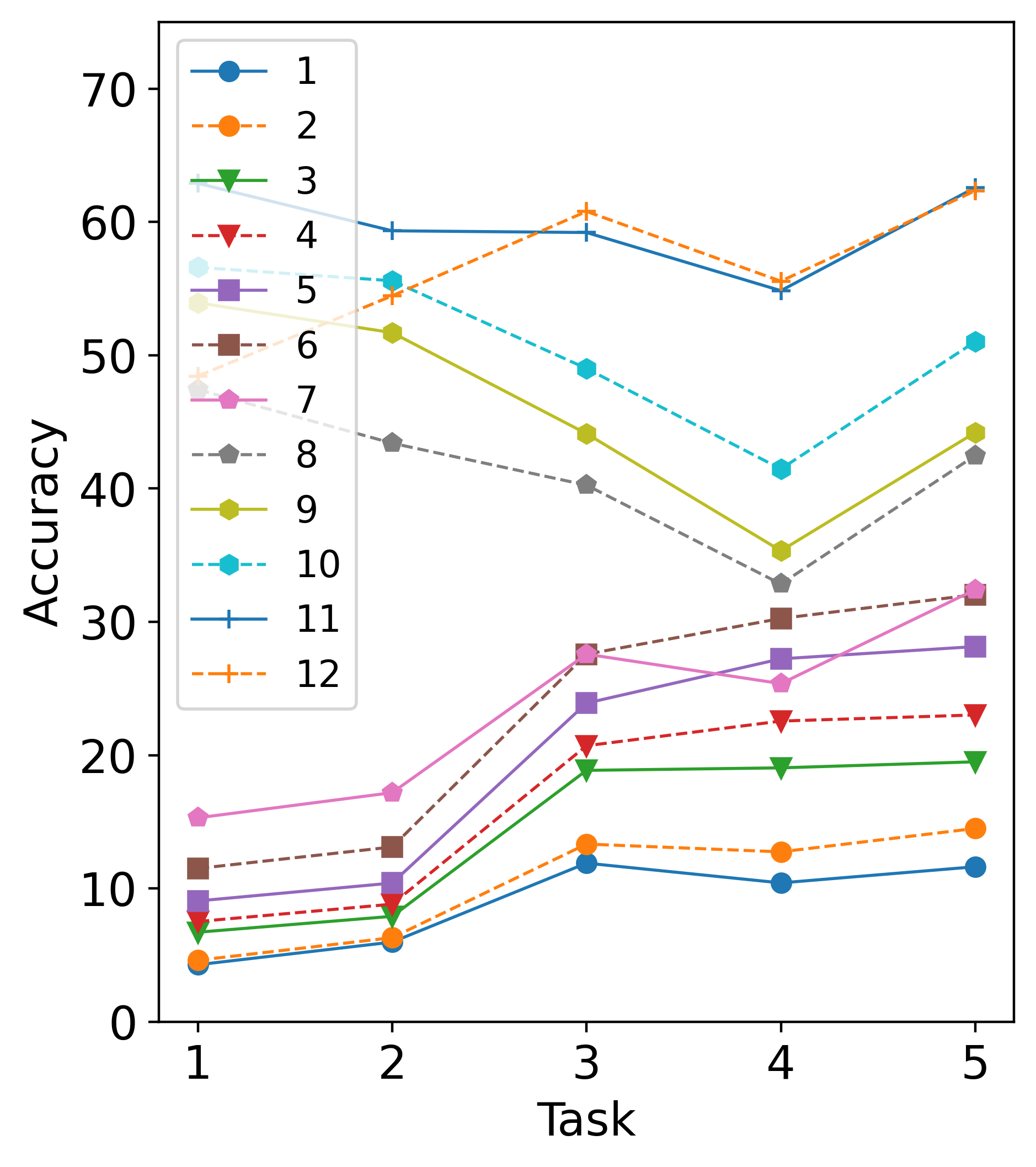} 
    \caption{ViLT + ER} 
\label{fig:task_probingb} 
\end{subfigure}%
\begin{subfigure}[b]{0.25\textwidth}
    \centering
    \includegraphics[scale=0.3]{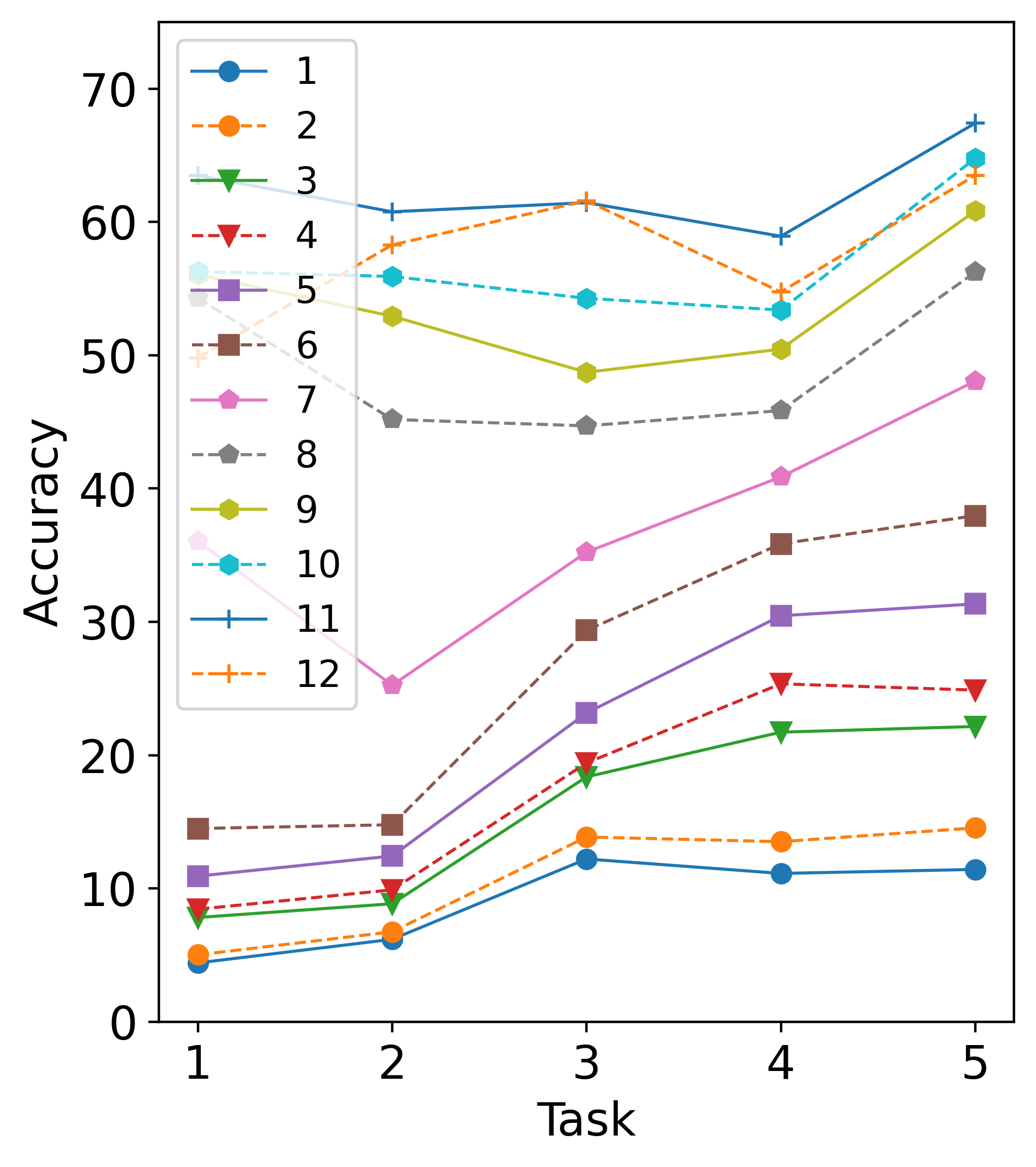} 
    \caption{ViLT + EWC} 
\label{fig:task_probingc} 
\end{subfigure}%
\begin{subfigure}[b]{0.245\textwidth}
    \centering
    \includegraphics[scale=0.3]{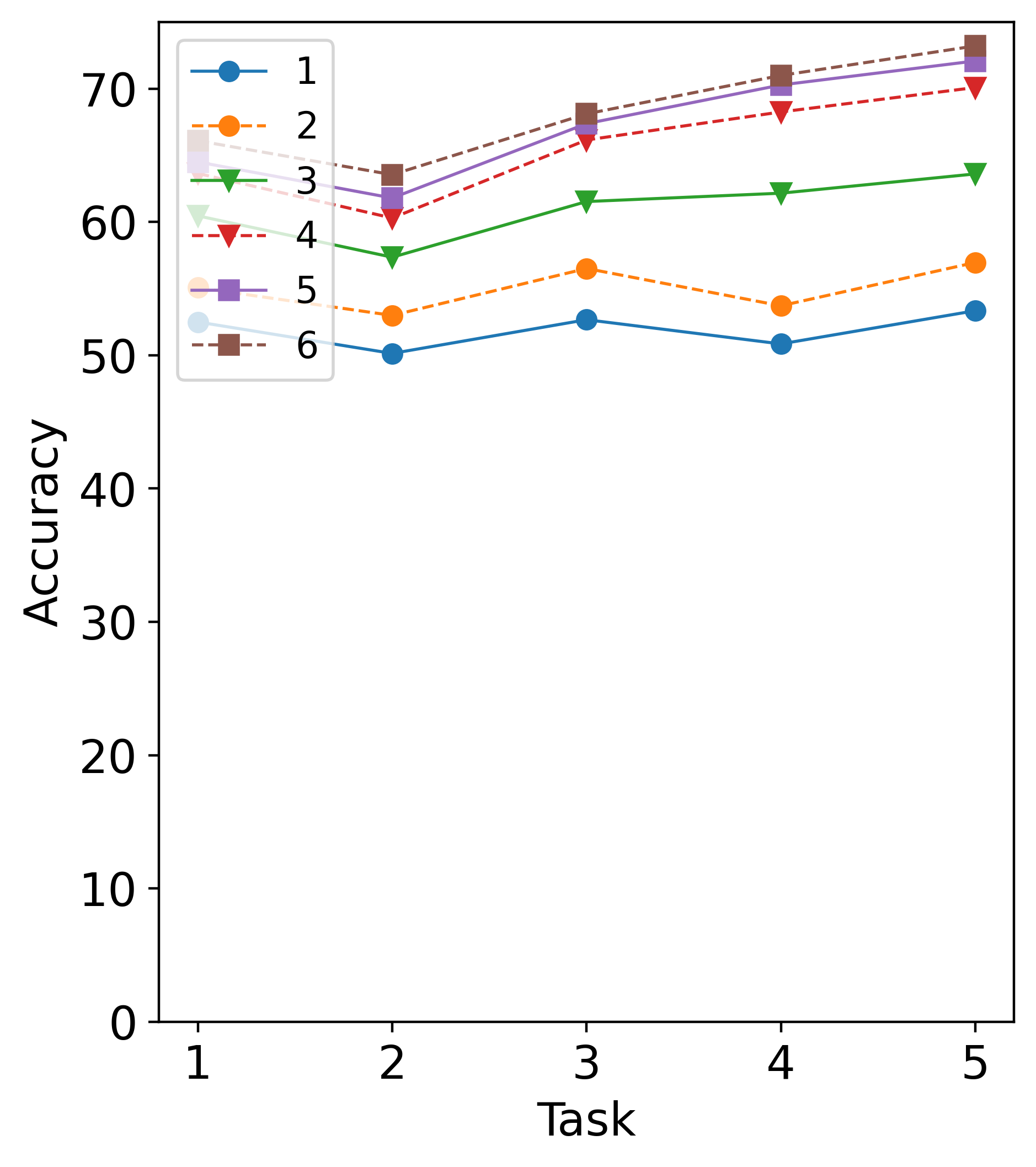} 
    \caption{ALBEF + ER} 
\label{fig:task_probingd} 
\end{subfigure}%
\caption{Task-wise probing results for ViLT and ALBEF model with different CL methods. The x-axis represents the model after training with task i, while the y-axis shows the \textit{representation accuracy}. (a-c): Line $i$ indicates the $i$-th layer of ViLT encoder. (d): Line $i$ indicates the $i$-th layer in the ALBEF multimodal encoder.}
\vspace{-15pt}
\label{fig:task_probing} 
\end{figure*}

\subsection{Impact of Vision and Text Encoder}
\label{sec:impact_vete}
In this section, we focus exclusively on vision and text encoders adopted in ALBEF. In order to further investigate the roles of the individual modules of ALBEF, we design the following experiments: 1) \textit{Freeze VE/TE}: we freeze the pre-trained vision encoder (or text encoder) and fine-tune the rest; 2) \textit{Freeze VE+TE}: we freeze both vision and text encoders, and fine-tune the rest; 3) \textit{Freeze B9}: besides vision and text encoder, we also freeze the bottom 3 layers of the multimodal encoder; 4) \textit{Freeze B12}: we freeze the full encoder, and only fine-tune the decoder. 

As shown in \cref{tab:albef_freeze}, the more modules are frozen, the less accurate the model is. 
\textit{Freeze B12} has the lowest accuracy, reflecting that due to its low plasticity, \textit{Freeze B12} prevents its feature extractor from adapting to new domains by learning new representations. On the other hand, thanks to its high stability, it suffers less from forgetting. 
Besides, its backward transfer performance indicates that the decoder also suffers from severe forgetting. Freezing the vision encoder or text encoder results in a similar performance to freezing the multimodal encoder. 

\subsection{Analysis of Multimodal Fusion Module}
\label{sec:mmfusion}
In this section, we investigate the impact of \textit{multimodal fusion module} on CL performance.

\subsubsection{Forgetting Analysis at Representation Level}
Following \cite{davari2022probing}, we first compute the average embedding for each class using the ground-truth label of all embeddings. Afterward, we re-classify all the embeddings by assigning the label of the closest class-wise embedding prototype and report the classification accuracy. We use the test set of each task and assume the ground-truth labels are available. Here, we provide the probing results for all 12 layers of ViLT as well as 6 layers in the multimodal encoder of ALBEF. We report the \textit{representation forgetting}, \ie, the average accuracy drop from the best performance on previous tasks, in \cref{fig:layer_prob_vilt}. Here, numbers smaller than zero mean there is positive knowledge transferred and benefits the performance of previous tasks. In \cref{fig:task_probing}, we provide the \textit{representation accuracy} of different methods and discuss our findings in the following section.

\begin{figure*}[ht]
\begin{subfigure}[b]{0.25\textwidth}
    \centering
    \includegraphics[scale=0.14]{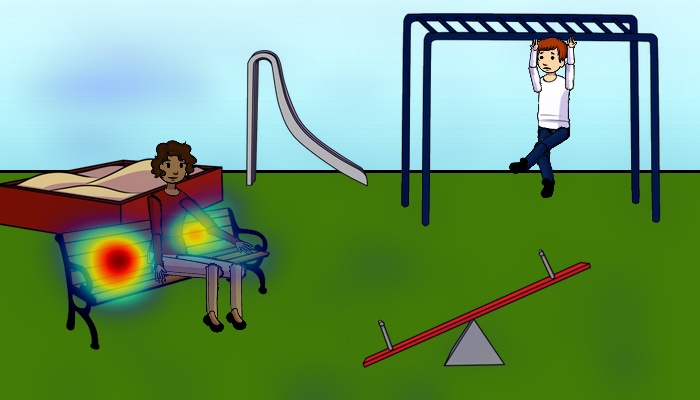} 
    \caption{Seq: Model after VQA Abstract} 
    \label{fig:grada}
\end{subfigure}%
\begin{subfigure}[b]{0.25\textwidth}
    \centering
    \includegraphics[scale=0.14]{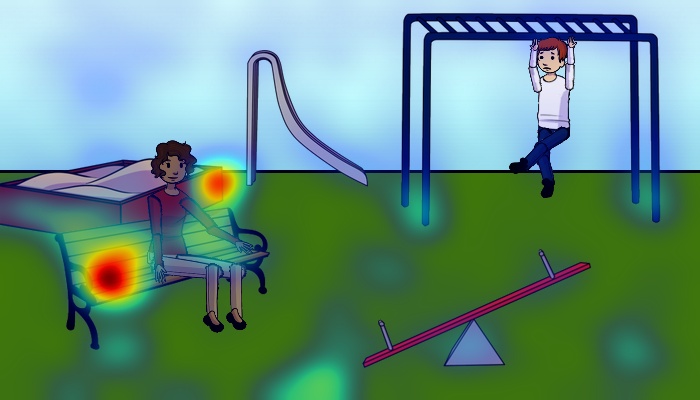} 
    \caption{Seq: Model after PathVQA} 
    \label{fig:gradb}
\end{subfigure}%
\begin{subfigure}[b]{0.25\textwidth}
    \centering
    \includegraphics[scale=0.14]{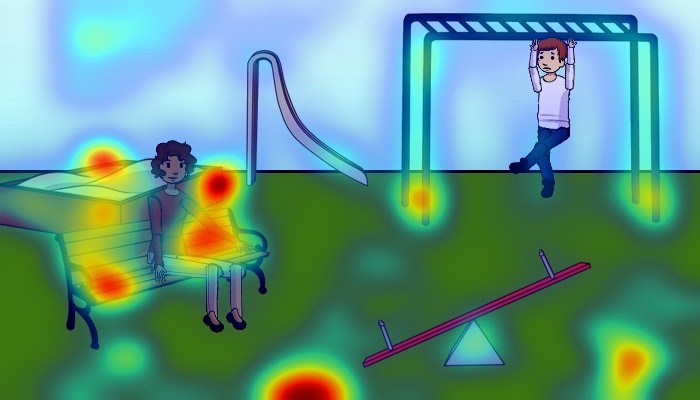} 
    \caption{Seq: Model after VQA-Med-2019} 
    \label{fig:gradc}
\end{subfigure}%
\begin{subfigure}[b]{0.245\textwidth}
    \centering
    \includegraphics[scale=0.14]{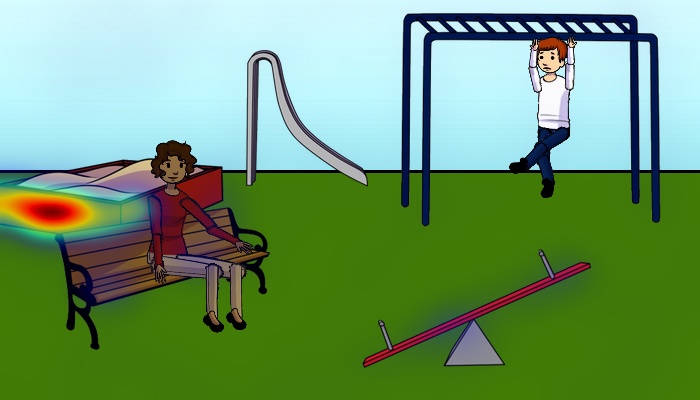} 
    \caption{Seq: Model after AQUA} 
    \label{fig:gradd}
\end{subfigure}%

\begin{subfigure}[b]{0.25\textwidth}
    \centering
    \includegraphics[scale=0.14]{./Figures/er_0.jpg} 
    \caption{ER: Model after VQA Abstract} 
    \label{fig:grade}
\end{subfigure}%
\begin{subfigure}[b]{0.25\textwidth}
    \centering
    \includegraphics[scale=0.14]{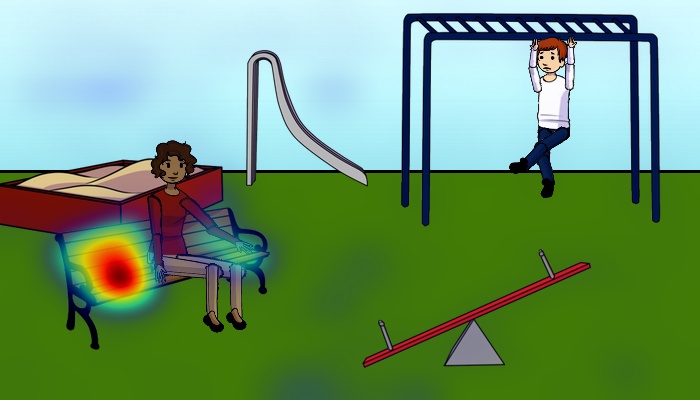} 
    \caption{ER: Model after PathVQA} 
    \label{fig:gradf}
\end{subfigure}%
\begin{subfigure}[b]{0.25\textwidth}
    \centering
    \includegraphics[scale=0.14]{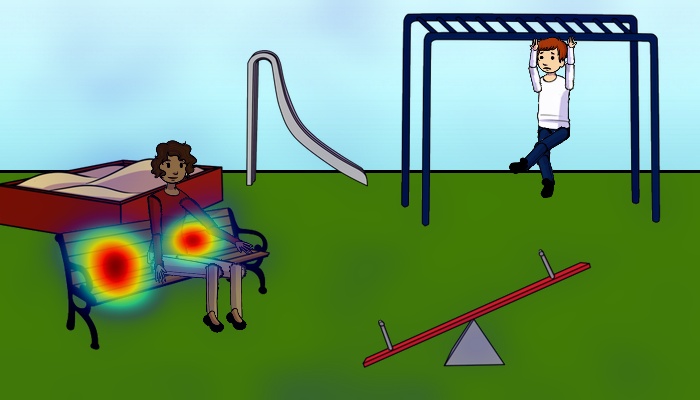} 
    \caption{ER: Model after VQA-Med-2019} 
    \label{fig:gradg}
\end{subfigure}%
\begin{subfigure}[b]{0.245\textwidth}
    \centering
    \includegraphics[scale=0.14]{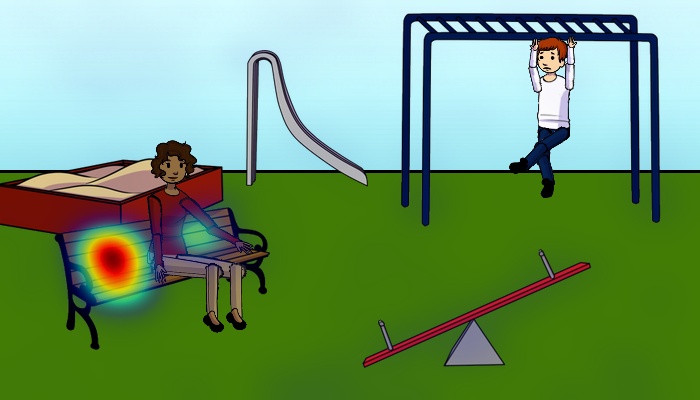} 
    \caption{ER: Model after AQUA} 
    \label{fig:gradh}
\end{subfigure}%
\caption{Grad-CAM visualization on the cross-attention maps of the multimodal encoder of ALBEF. Seq: sequential fine-tuning baseline. The task order is shown in \cref{fig:main_illu}. Question: What color is the bench? Ground truth answer: Brown. }
\vspace{-10pt}
\label{fig:grad} 
\end{figure*}

\subsubsection{Results and Discussions}
\paragraph{The deeper layers of the single-stream model are most affected by forgetting.}
For single-stream architecture ViLT, the amount of representation forgetting increases with the increase of layers (see \cref{fig:layer_prob_vilta}). For the lower layers, the representation forgetting stays relatively flat or decreases slightly, showing that the bottom layers barely suffer from forgetting and a positive backward transfer is occurring. However, the representation forgetting increases significantly at layer 7, and then the deeper the layer, the more representation forgetting we observe. This phenomenon may be due to the fact that in the lower layers, the model only learns general knowledge. However, the model tends to learn more task-specific features in the upper layers, which are easily forgotten and corrupted when constantly learning new tasks. Moreover, we find that the representation accuracy of the last layer is lower than the second last layer (see \cref{fig:task_probinga} and \cref{fig:task_probingc}), suggesting that applying another classification method, \eg, metric learning, on the representations extracted from the second to last layer might be a more viable approach to adapt the single-stream models to the CL environment. We assume that this is caused by the introduction of a task-specific classifier, which harms the representation ability of the final layer embedding when switching between tasks.

\paragraph{ER helps align the single-stream encoder with the task-specific classifier.}
Comparing \cref{fig:layer_prob_vilta} with \cref{fig:layer_prob_viltb}, we notice that the representation forgetting of ER is similar to that of the sequential fine-tuning baseline, however, ER achieves much higher backward transfer than the baseline (see \cref{tab:result_main}). We believe that ER and other replay-based approaches may assist in aligning the encoder with the task-specific classifier. In particular, it is worth noting that, although the forgetting in terms of the traditional measure is high for EWC (see \cref{tab:result_main}), the representation forgetting shown in \cref{fig:layer_prob_viltc} indicates that EWC is more effective at mitigating forgetting of the middle layers. However, it lacks the ability to constrain the classifier when a task-specific classifier needs to be utilized.

\paragraph{Fusion module of a dual-stream model forgets less than that of a single-stream model.}
Comparing \cref{fig:layer_prob_viltb} \cref{fig:layer_prob_viltc} with \cref{fig:layer_prob_vilte} and \cref{fig:layer_prob_viltf}, we observe that ALBEF shows much milder representation forgetting than ViLT, which is also evidenced by the best CL performance achieved by ALBEF (see \cref{tab:result_main}). Besides, we notice that across all CL methods, the representativeness of the layers within the multimodal encoder of ALBEF does not vary as dramatically as ViLT, possibly because the forgetting of ALBEF occurs mainly in the vision encoder, text encoder, and the decoder, which is also evidenced by \cref{tab:albef_freeze}. Also, as shown in \cref{fig:task_probingd}, the multimodal encoder of ALBEF extracts representations with higher separability.

\subsection{Visualization}
\label{sec:vis}
\cref{fig:grad} shows the Grad-CAM \cite{DBLP:journals/corr/SelvarajuDVCPB16} visualization changes of the continually adapted VLPM on a selected image from the first task. Following \cite{li2021align}, we compute the Grad-CAM on the cross-attention maps in the 3rd layer of the multimodel encoder in ALBEF for the VQA task. From the first row in \cref{fig:grad}, we see that the model attention shifts from the target object to almost every possible object existing in the image after being trained on 2 medical datasets, which we assume are due to the large domain shift in the visual aspect. \cref{fig:gradd} shows that the model attention again focuses on a certain object but fails to target the correct one, which is a predictable behavior since, in the AQUA dataset, objects are also abstractions of real-life objects. The second row of \cref{fig:grad} shows the same Grad-CAM visualization of ALBEF optimized with ER. We notice that by replaying instances from the VQA Abstract dataset, the model maintains the ability to distinguish abstract objects and connect them with the questions while learning new tasks.

\section{Conclusion}
\label{sec:conclu}
In this paper, we present CL-CrossVQA, the first VQA benchmark for cross-domain continual learning. We systematically investigate which type of CL method is the most effective for VLPMs, and how model design impacts CL performance. Our comprehensive experimental results reveal that: 1) Dual-stream encoder-decoder model exhibits alleviation of forgetting; (2) The deeper layers of single-stream encoder-only models are most affected by forgetting; 3) Replay-based approaches are the most effective and are less sensitive to task orders; 4) Adapter is a promising approach in this cross-domain continual learning scenario. We believe that the insights gained by dissecting the inner architecture of VLPMs will inform the development of CL approaches suitable for multimodal continual learning.

{\small
\bibliographystyle{ieee_fullname}
\bibliography{CL_CrossVQA}
}

\end{document}